\documentclass{acm_proc_article-sp}
\usepackage{graphicx,subcaption, amsmath, hyperref, dsfont}

\begin{document}

\conferenceinfo{}{Bloomberg Data for Good Exchange 2016, NY, USA}

\title{Quantifying urban traffic anomalies}

\numberofauthors{3}
\author{
\alignauthor
Zhengyi Zhou\\
       \affaddr{AT\&T Labs Research}\\
       \affaddr{New York, NY}\\
       \email{\normalsize{zzhou@research.att.com}}
\alignauthor
Philipp Meerkamp\\
       \affaddr{Bloomberg LP}\\
       \affaddr{New York, NY}\\
       \email{\normalsize{pmeerkamp@bloomberg.net}}
\alignauthor 
Chris Volinsky\\
       \affaddr{AT\&T Labs Research}\\
       \affaddr{Bedminster, NJ}\\
       \email{\normalsize{volinsky@research.att.com}}
}

\maketitle

\begin{abstract}
Detecting and quantifying anomalies in urban traffic is critical for real-time alerting or re-routing in the short run and urban planning in the long run. We describe a two-step framework that achieves these two goals in a robust, fast, online, and unsupervised manner. First, we adapt stable principal component pursuit to detect anomalies for each road segment. 
This allows us to pinpoint traffic anomalies early and precisely in space. 
Then we group the road-level anomalies across time and space into meaningful anomaly events using a simple graph expansion procedure. These events can be easily clustered, visualized, and analyzed by urban planners. 
We demonstrate the effectiveness of our system using 7 weeks of anonymized and aggregated cellular location data in Dallas--Fort Worth. 
We suggest potential opportunities for urban planners and policy makers to use our methodology to make informed changes. These applications include real-time re-routing of traffic in response to abnormally high traffic, or identifying  candidates for high-impact infrastructure projects. 
\end{abstract}

\section{Introduction} \label{sec:intro}

High and sub-optimally flowing road traffic are prevalent in US urban areas, and have severe negative consequences related to
\vspace{-2mm}
\begin{itemize}
\item Time and money: congestion results in an average loss of 42 hours a year per commuter, and \$160 billion per year in the US \cite{UTAM}.
\item Public health: long commutes have been linked to heart disease, high blood pressure, obesity and stress \cite{Hoehner:2012, Hansson:2011}. 
\item Environment: congested roads tend to reduce the fuel efficiency of cars, resulting in increased emissions of carbon dioxide and pollutants. The study in \cite{Barth:2008} estimates that traffic-related emission reductions of up to 20\% could be achieved solely by enabling a smoother traffic flow. 
\item Emergency response: congestion can slow down dispatch of emergency vehicles responding to medical, fire, and security emergencies. For example, \cite{Larsen:1993} estimate that every minute of delay in emergency response to cardiac arrest reduces the patient's probability of survival by 7-10\%. 
\end{itemize}
\vspace{-2mm}
We propose a methodology to identify ``expected'' patterns in urban traffic, and detect traffic anomalies, or deviations from the expected. Anomalies can be caused e.g. by accidents, road works, weather, or events. We envision two main use cases of our methodology. 

First, in the short run, real-time detection of traffic anomalies at the road level can enable timely alerting and re-routing of traffic, e.g. through push notifications, radio announcements, or various message boards. This can prevent traffic build-up, wasted time, and additional pollution; it allows for proactive planning, which may be especially important for emergency response vehicles.

Second, in the long run, quantifications of expected traffic flow and anomalies can help urban planners monitor, analyze, and modify traffic systems. For instance, urban planners can use these quantifications to identify high-impact infrastructure projects, e.g., removing bottle necks in the road system, evaluating public transportation capacity and routes, minimizing impact of road work, or optimizing traffic light scheduling. They can also use these quantifications to justify projects or changes that would be too expensive to implement without detailed prior evaluation. 

As another example, cities can better prepare for major urban events, such as sports games or concerts. Insights can be obtained on where and when attendees come from or leave after such events, or how to foresee and proactively alleviate traffic problems, e.g., through public transportation or ride sharing. This paper highlights, as an example, the impact of a major football game on traffic flow the city Dallas. Our analysis and data provide a basis for urban planners to develop traffic management strategies for similar events.

We address both the short-term goal of real-time re-routing and the long-term goal of facilitating urban planning. To this end, we propose a two-step framework: (i) detect traffic anomalies efficiently and robustly at each road segment using stable principal component pursuit, to pinpoint anomalies early on (Section \ref{sec:spcp}); (ii) group the road-level anomalies across time and roads into meaningful events using graph expansion. These events can be easily clustered, visualized, and analyzed by urban planners (Section \ref{sec:group}). Our approach is
\vspace{-2mm}
\begin{itemize}
\item Spatially precise: we can identify anomalies at the road level, while most prior studies operate on on large regions or grids (see Related work in Section \ref{sec:lit}).
\item Robust: we can accurately detect anomalies that are loosely correlated across time using heterogeneous traffic data that do not follow simple parametric assumptions (more on data challenges in Section \ref{sec:data}).
\item Simple and fast: the two-step pipeline is efficient, online, unsupervised, and gives results interpretable by and friendly to non-experts such as urban planners.
\end{itemize}
\vspace{-2mm}

\subsection{Data and challenges} \label{sec:data}

Urban planners have traditionally relied on surveys, traffic cameras, or under-pavement road sensors, which are expensive to install and maintain throughout cities. The increasing availability of mobile devices provides an unprecedented opportunity to measure and analyze urban mobility. 

\begin{figure}[h]
\centering
\hspace*{-0.3cm}
\includegraphics[width=1.05\linewidth]{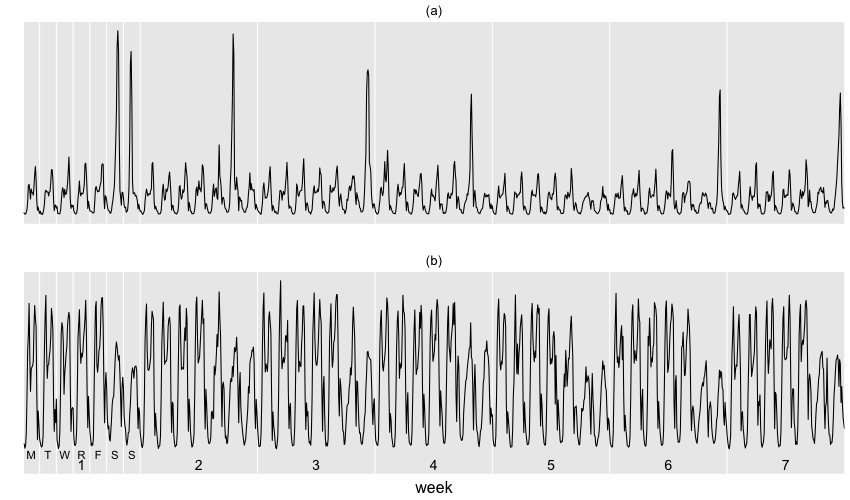}
\caption{Traffic volumes from two sample roads: (a) a road near AT\&T Stadium, with 7 spikes in traffic volumes corresponding to 6 football games and a concert, mostly occurred on weekend evenings; (b) a highway, with no visually obvious traffic anomalies. We show in Figure \ref{fig:anomaly} the decompositions of traffic for these two roads into the expected and the anomalous using method described in Section \ref{sec:spcp}.}
\label{fig:ts}
\end{figure}

We use anonymized and aggregated cellular data to infer traffic flow in Dallas-Fort Worth, Texas for a 7-week period in late 2015. Cellular network positioning has the advantage of lower battery consumption and higher coverage compared to GPS \cite{Perera:2015}, but the location samples can be quite noisy; we only observe devices at a relatively sparse set of cell tower locations connected to the devices. The top set of roads these devices could be traveling on are inferred using  a probabilistic matching algorithm \cite{Raif}. The traffic volumes on most roads exhibit strong weekly seasonality (see Figure \ref{fig:ts}). 

The road network in Dallas-Fort Worth (and many other cities) is highly heterogeneous. Roads close-by (e.g., highway and an exit), or the same road in different directions can have very different traffic behaviors.

Most urban traffic anomalies do not occur completely randomly in time and space, but with some autocorrelation or loose regularity. For instance, big concerts or sport events typically occur on weekend evenings near major stadiums (see Figure \ref{fig:ts} (a)); traffic may be impacted for several hours prior or after these events; major congestions are more likely to happen during rush hours on weekdays along popular commuting routes.

\subsection{Related work} \label{sec:lit}
Many traffic anomaly detection methods are region-based, discretizing a city into several regions, and anomaly detection is done across regions. For example, \cite{Liu:2011} and \cite{Chawla:2012} partition the city into disjoint regions by major roads, and find unexpected traffic between any two regions. Studies \cite{Pang:2011} and \cite{Pang:2013} partition a city into uniform grids and report anomalies if neighboring cells have very different total traffic volumes. However these region-based methods cannot identify anomalies at the road level, are sensitive to how regions are defined (coarse partition may result in loss of information), and suffer from a ``boundary'' problem (traffic or anomaly on the boundary may be dispersed into multiple regions); these downsides are described in detail in \cite{Lan:2014} and \cite{Doraiswamy:2014}.  

The existing road-based traffic anomaly detection methods make strong assumptions on the traffic. For example, \cite{Lan:2014} and \cite{Shekhar:2001} assume that historical or neighborhood traffic is normally distributed, and compare current traffic to history or neighborhood using the mean and standard deviation parameters of fitted normal distributions. We found that traffic in most roads of Dallas, even if restricted to similar times across weeks (e.g., every Monday 8-9am) or a small neighborhood, does not follow normal distributions, but can be light-tailed, heavy-tailed, or multi-modal, partially due to the lower bound at 0 and somewhat recurring anomalies. 

Studies that use spatial similarity such as \cite{Doraiswamy:2014} and \cite{Shekhar:2001} are inappropriate for highly heterogeneous road network that most cities have. 
Most studies use GPS locations of taxis. Our dataset is not restricted to taxis, and may portray urban traffic more realistically. 

\section{Detecting anomalies for each road} \label{sec:spcp}
We aim to detect anomalies in traffic volumes separately for each road in Dallas in a robust, fast, online, and unsupervised manner. We use an approach based on a variant of principal component analysis (PCA) that is robust to outliers and noise.

\subsection{Stable Principal Component Pursuit}
PCA has been used for anomaly detection in many applications, \cite{Jolliffe:2002}, , in particular for (city-wide) traffic across regions \cite{Chawla:2012} and (network-wide) network traffic \cite{Lakhina:2004}.  
but studies have shown that it is not robust to outliers and noise 
\cite{Ringberg:2007},  and 
an effective way to address this is robust PCA (RPCA) solved via principal component pursuit (PCP) \cite{Candes:2011, Wright:2009}.  PCP decomposes a data matrix $T$ as $T = L + A$, where $L$ is low-rank fit of $T$ and $A$ is sparse matrix representing anomaly, i.e. the aim is to solve
\begin{equation}
\min_{L,A}  \,\, \mathrm{rank}(L) + \lambda ||A||_0	, \,\,\,\,\,\,\, \mbox{subj} \,\,\,\,\,\,\,\, T - L - A = 0, \label{eqn:rpca}
\end{equation}
where $\lambda>0$ is a linear combination parameter, usually taken to be $1/\sqrt{\mbox{max}(m,n)}$, where the data matrix is $m \times n$. 
Here, 
$||.||_0$ is the $\ell_0$-norm that encourages entry-wise sparsity. The  minimization in (\ref{eqn:rpca}) is NP-hard, and we solve instead its convex relaxation: 
\begin{equation}
\min_{L,A}  \,\, ||L||_* + \lambda ||A||_1	, \,\,\,\,\,\,\, \mbox{subj} \,\,\,\,\,\,\,\, T - L - A = 0, \label{eqn:pcp}
\end{equation}
where $||.||_*$ is the nuclear norm ($\ell_1$-norm of singular values) and $||.||_1$ is the $\ell_1$-norm. 

RPCA-PCP assumes $L$ to be exactly low-rank, $A$ to be exactly sparse, and data to be exactly the sum of the two. However, data in many applications is often corrupted by random and systematic noise affecting many entries of the data matrix. A relaxed PCP,  Stable PCP (SPCP)  \cite{Zhou:2010}, incorporates entry-wise noise by decomposing $T = L + A + E$ with an extra error term $E$ and constraining its Frobenius norm ($||.||_F$, entry-wise $\ell_2$-norm) to be small, i.e.,
\begin{equation}
\min_{L,A}  \,\, ||L||_* + \lambda ||A||_1	, \,\,\,\,\,\,\, \mbox{subj} \,\,\,\,\,\,\,\, ||T - L - A||_F < \delta, \label{eqn:spcp}
\end{equation}
for some $\delta>0$. RPCA-PCP and SPCP have been successfully applied to applications in vision and compressive sensing, e.g., video surveillance \cite{Bouwmans:2014} and face recognition \cite{Candes:2011}. 

\subsection{Adapting SPCP for traffic anomalies}
We begin with an hourly time series of traffic volume for each road. We convert each week of hourly traffic volume into a column vector and stack each week's vector to form a data matrix $T$. We exploit the strong weekly seasonality observed in data, and assume, except for occasional anomalies, the weekly pattern of traffic volumes to be fairly stable. Intuitively, we assume that the original traffic volume ($T$) is the sum of three terms: (i) the ``expected'' city traffic modeled by a low-rank subspace that can gradually change over time ($L$), (ii) traffic anomalies modeled by sparse and potentially correlated outliers ($A$), and (iii) noise and uncertainty in the 
location samples and trajectories 
($E$).

\begin{figure}[h]
\centering
\hspace*{-0.3cm}
\includegraphics[width=1.05\linewidth]{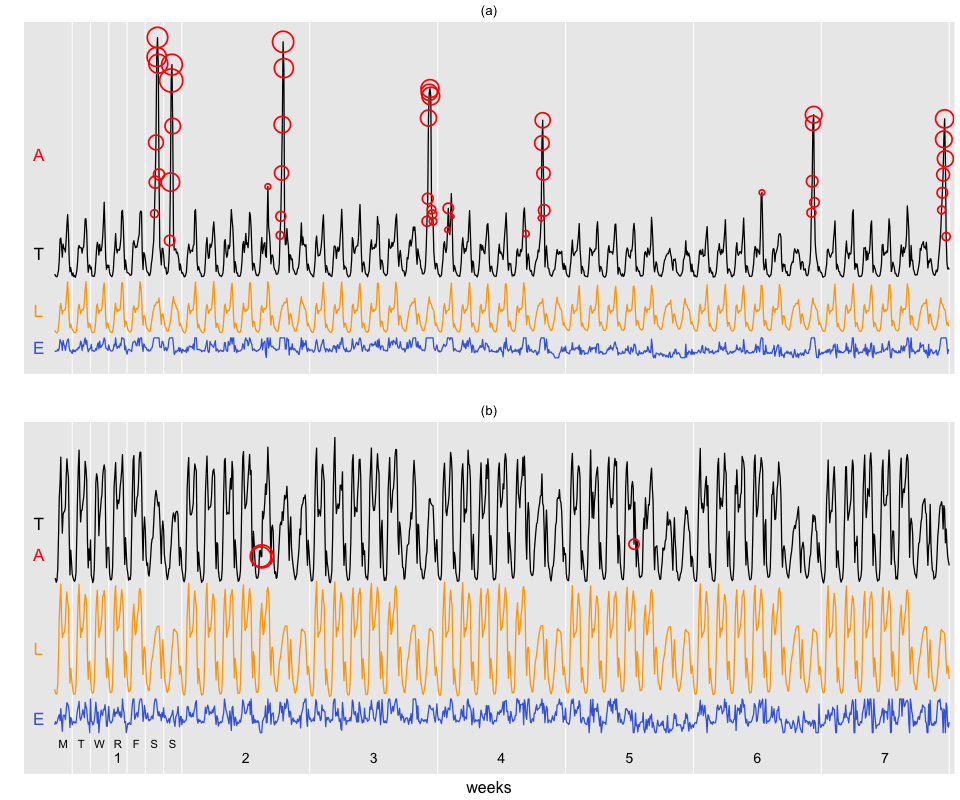}
\caption{SPCP decomposition of traffic on the two roads shown in Figure \ref{fig:ts}. The original traffic volume time series $T$ (black, top) is decomposed into the sum of low-rank fit $L$ (orange, middle), noise $E$ (blue, bottom), and sparse anomalies $A$ (non-zero entries in red circles with size corresponding to absolute values). (a) a road near the AT\&T Stadium, showing the impact of 6 football games and a concert; (b) a highway, showing an unidentified reduction in a Friday morning rush in week 2 and the impact of a heavy thunderstorm in week 5.}
\label{fig:anomaly}
\end{figure}

Figure \ref{fig:anomaly} shows the SPCP decomposition of traffic on the two roads shown in Figure \ref{fig:ts}. For each road, we obtain a low-rank fit $L$ that represents the weekly pattern that may differ slightly from week to week. We found $L$ to have rank one or two for most roads (two for road (a) and one for (b) in Figure~\ref{fig:anomaly}). The anomaly matrix $A$ is sparse and successfully captures the impact form a variety of events. For (a), we detected all 7 sport and concert events, even though these anomalies are correlated across hours (persisting for several hours) and weeks (mostly occurring on weekend evenings). For (b), we detected an unrealized Friday morning peak commute in week 2 (reason unknown) and a dip in week 5 around the onset of a heavy thunderstorm. 

After the SPCP decomposition, we normalize anomaly $A$ to get $\tilde A = A / L$ element-wise, i.e. compare anomalies with the expected traffic volume, for meaningful comparison across time and roads. The values of $\tilde A$ are interpretable: positive for higher than expected, negative for lower, with the numerical magnitude corresponding to the amount of deviation from expectation. This allows for easy filtering, thresholding, and comparisons across regions and times.  

For road and time combinations whose median traffic volume across weeks is too low, we override their $\tilde{A}$ values to NA (the ratio is unstable when $L$ is close to zero). In practice, these road-time combinations with sparse traffic volumes could be combined with neighboring roads or examined at coarser grids in time.

\begin{figure*}[h]
\centering
	\begin{subfigure}{0.317\linewidth}
	\centering\caption{}
	\includegraphics[width = \linewidth, height = 0.85\linewidth]{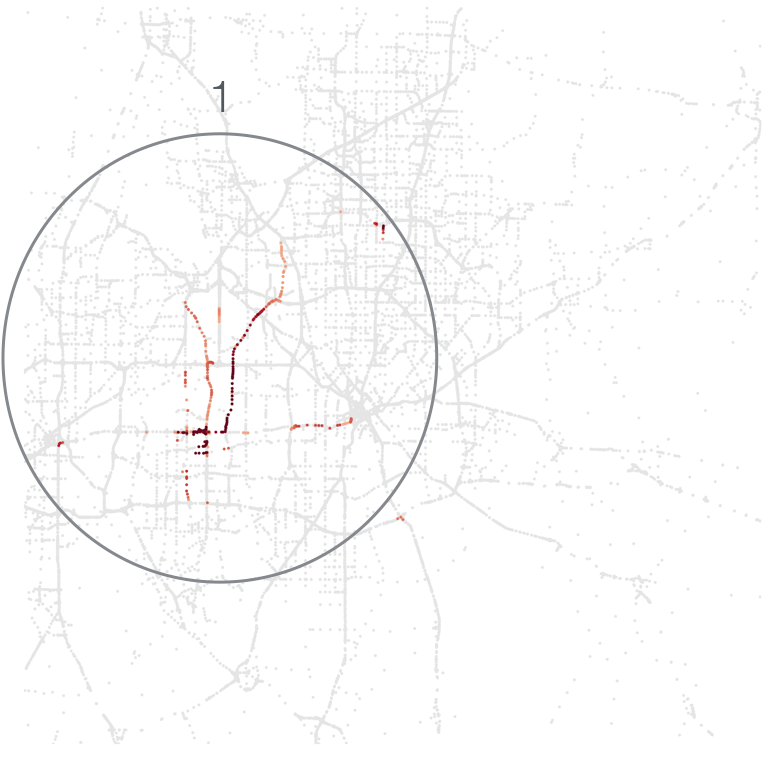}
	\end{subfigure}
	\begin{subfigure}{0.317\linewidth}
	\centering\caption{}
	\includegraphics[width = \linewidth, height = 0.85\linewidth]{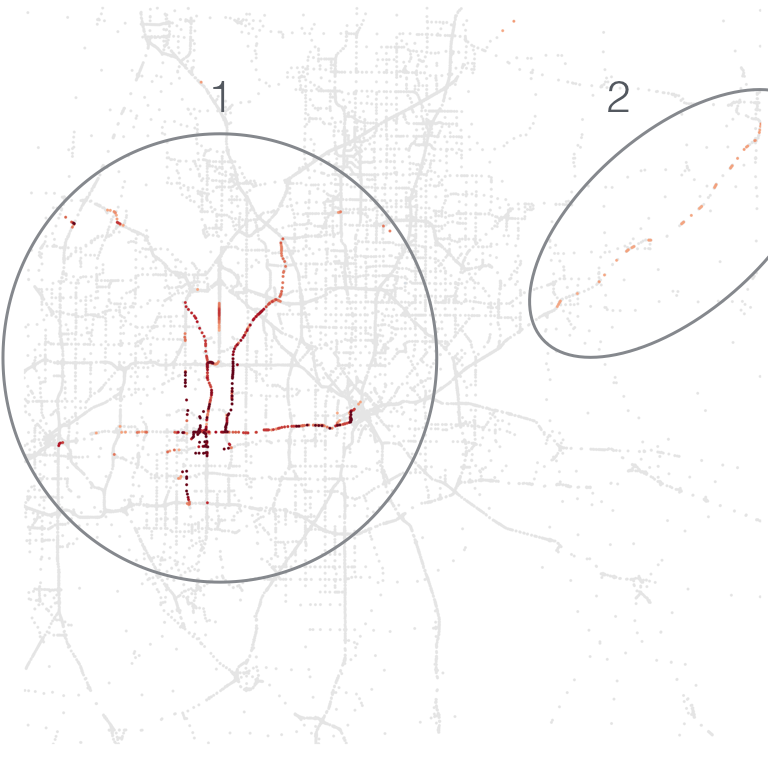}
	\end{subfigure}	
	\begin{subfigure}{0.317\linewidth}
	\centering\caption{}
	\includegraphics[width = \linewidth, height = 0.85\linewidth]{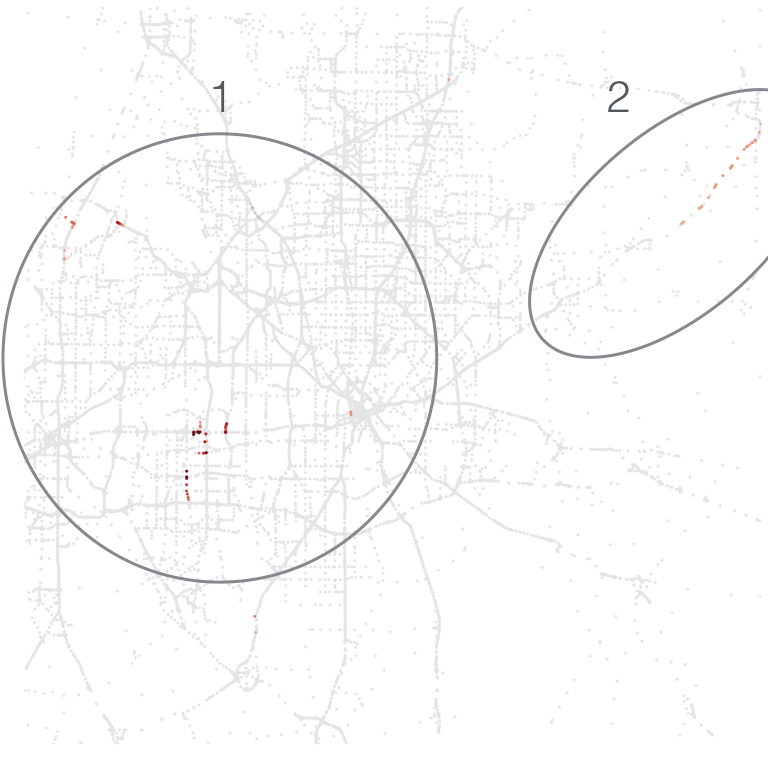}
	\end{subfigure}		
	\begin{subfigure}{0.03\linewidth}
	\centering
	\includegraphics[width = \linewidth, height = 1.8in]{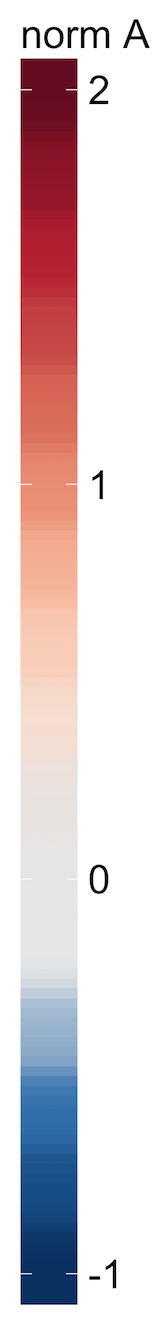}
	\end{subfigure}
	\begin{subfigure}{0.317\linewidth}
	\centering\caption{}
	\includegraphics[width = \linewidth, height = 0.85\linewidth]{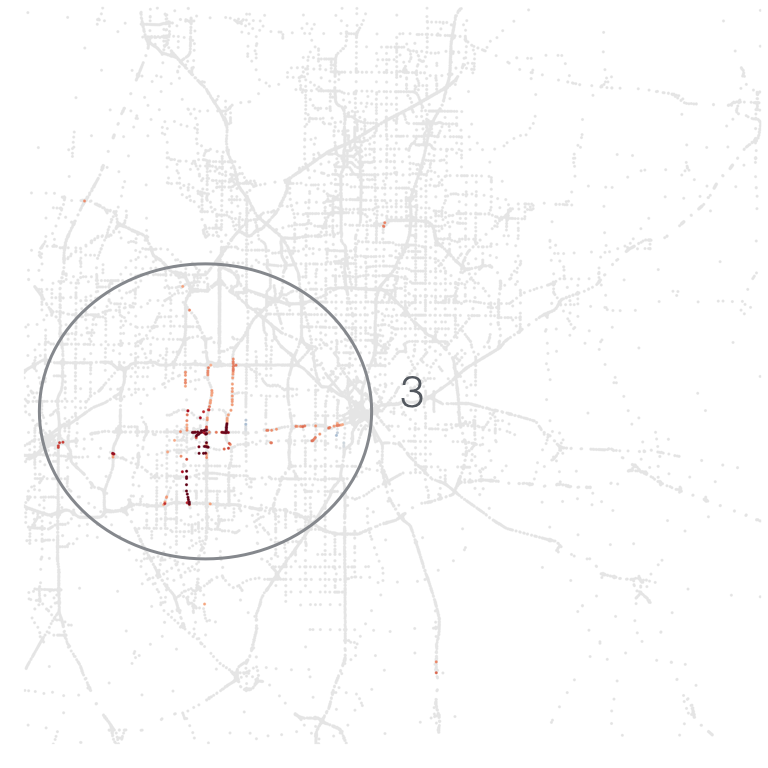}
	\end{subfigure}
	\begin{subfigure}{0.317\linewidth}
	\centering\caption{}
	\includegraphics[width = \linewidth, height =0.85\linewidth]{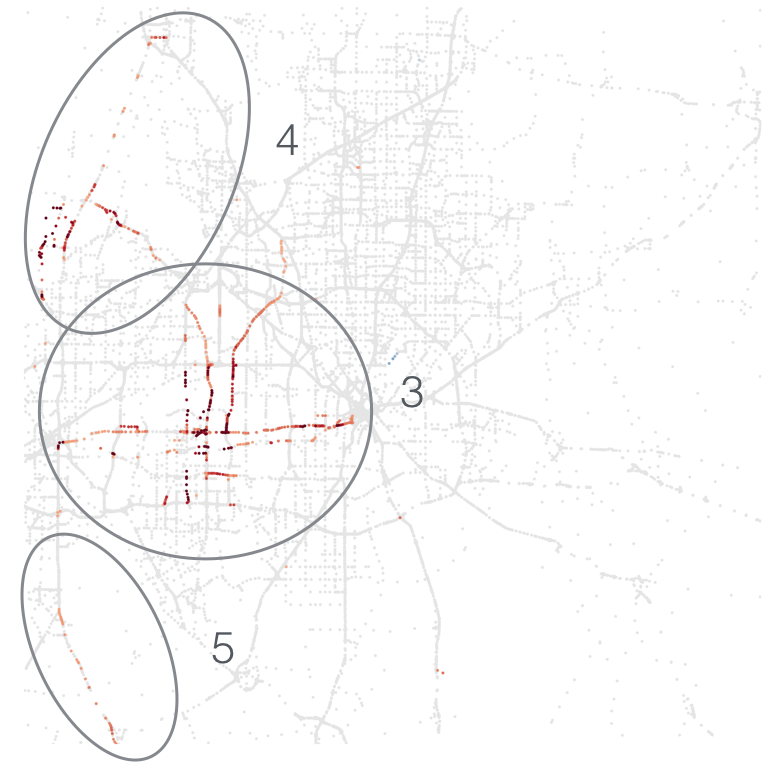}
	\end{subfigure}
	\begin{subfigure}{0.317\linewidth}
	\centering\caption{}
	\includegraphics[width = \linewidth, height = 0.85\linewidth]{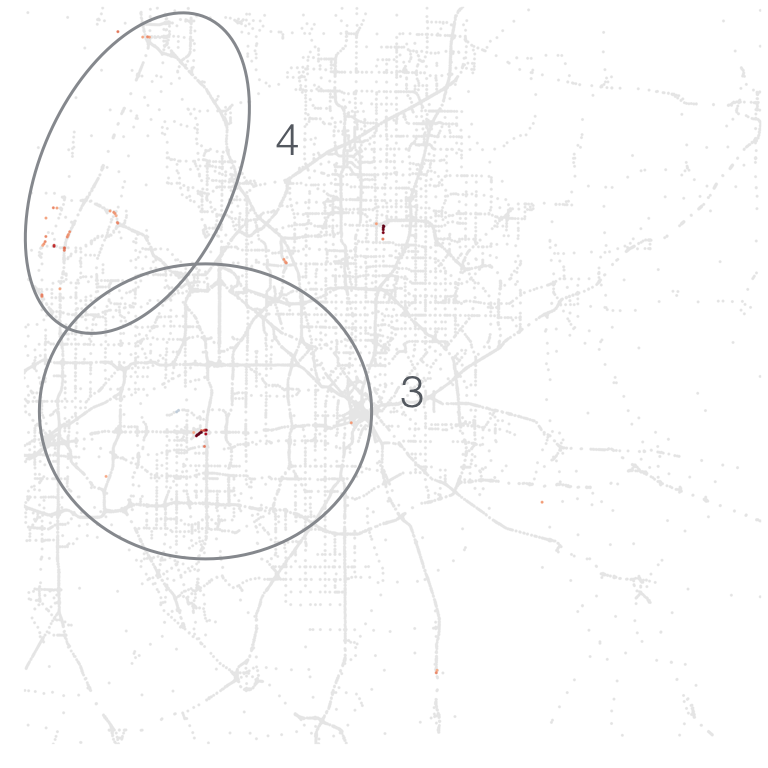}
	\end{subfigure}
	\begin{subfigure}{0.03\linewidth}
	\centering
	\includegraphics[width = \linewidth, height = 1.8in]{legend}
	\end{subfigure}
	\caption{The impact of a major NFL game on the city: (a) - (c) show the traffic surges in the three hours leading up to the game as attendees traveled to the AT\&T Stadium, and (d) - (f) for the three hours after the game as attendees left or pursued post-game activities. The midpoint of each road is plotted in a color corresponding to its normalized anomaly value $\tilde A$. The ovals represent the evolution of 5 anomaly events, obtained by grouping road-level anomalies across time and space using method described in Section \ref{sec:group}.}
	\label{fig:nfl}
\end{figure*}

We show in Figure \ref{fig:nfl} the impact of a major NFL football game on the city.  The midpoint of each road is colored according to its normalized anomaly value $\tilde A$. The top panel (a) - (c) shows the increase in traffic in the three hours before the game as the attendees traveled towards the AT\&T Stadium. The bottom panel (d) - (f) shows the increase in traffic three hours after the game, as the attendees left the stadium or pursued post-game activities. There were no major anomalies during the game. It is interesting that, on the aggregate, attendees arrive quite early to the game and via routes  perhaps different from those they took to leave. 

It is hard to get accurate ground-truths for urban traffic anomalies, both anomaly labels and root cause. We have examined many events in the city for which we have rough positions in time and space, including football, basketball, hockey, and baseball games, concerts, and severe weather. For each event, we obtained reasonable and highly intuitive results like those in Figure \ref{fig:nfl}, confirming the effectiveness and robustness of our approach. Our method also found many other anomalies, which we were unable to associate with particular events, but could plausibly correspond to traffic congestion, blockage, or rerouting. More generally, the lack of ground truth for traffic anomalies precludes supervised or semi-supervised methods, and presents a central challenge in detecting anomalies.

\subsection{Computation and alternative approaches}
There are many fast algorithms for solving SCPC \cite{Aybat:2014, Aravkin:2014, Tao:2011}, with open-source code available for many of them. We use the Partially Smooth Proximal Gradient method proposed in \cite{Aybat:2014}.  Many real-time implementations are also proposed, and we adapt the ``seeding'' idea from \cite{Pope:2011}, where $L$ and $E$ are updated relatively infrequently, and their old values are used as a starting point for the update. In our application, we found updating $L$ and $E$ once a week yields almost identical results as updating them hourly. The computation is embarrassingly parallelizable across roads, and computing for the entire city for the 7-week period took a small number of hours on a personal computer. 

There are many variants of SPCP, e.g., using the $\ell_\infty$-norm or Huber penalty for the error matrix $E$, or using the row-wise $\ell_2$-norm sum for the anomaly matrix $A$ \cite{becker:2011, Aravkin:2014, Mateos:2012}. There are also other approaches to solve RPCA, including Bayesian RPCA \cite{Ding:2011, Babacan:2011} and Approximated RPCA \cite{Zhou:2011}. Most of these approaches have been shown to give similar results \cite{Bouwmans:2014}. 

Other than SPCP and RPCA-PCP methods, we also considered a wide range of other approaches, but found most of them not to be robust enough. For example, seasonality methods based on Fourier, wavelet, seasonal autoregressive moving average models \cite{Box:2008}, or seasonal and trend decomposition using LOESS \cite{Cleveland:1990} tend not to work well if there are more than one large outliers occurring almost in sync with the seasonality, which we observe in our data not infrequently. Some of these seasonality methods also use square loss in the estimation, which undesirably amplifies the impact of outliers. We also considered approaches based on robust measures, e.g., median, absolute value, median absolute deviation, or Huber weights \cite{Huber:2004}. We found cases where correlations of anomalies across hours and weeks corrupt even these robust methods, unless some supervision is given to annotate and exclude most of the anomalies. 

One could also use more sophisticated methods such as Gaussian processes or hidden Markov models, but the high degree of noise and loosely correlated outliers in the data requires robust methods. Combining a robust approach like ours with these sophisticated methods may be beneficial.

\section{Grouping anomalies} \label{sec:group}
Given anomaly measures on all road segments, we can group anomalies across time and space into anomaly events. We propose a simple graph expansion procedure to accomplish this.

Starting from any anomalous road segment as the seed, we iteratively append all anomalous road segments that are within $n$~degrees of neighbor on the road network and within $s$~time periods before and after. This process is continued via either breadth- or depth-first-search until there is no more anomaly in the vicinity in time and space. We group together a subgraph of road, time combinations in the space of road network $\times$ discrete time, and call this an anomaly event.  We then choose any ungrouped anomalous road segment as the new seed and repeat this procedure until no anomalous road segment remains. 

This process is invariant to the choice of seed and the order of appending. The degree of neighbor parameters $n$ and $s$ can be fixed, or allowed to vary across space (e.g., by regions in city, length or type of road) or time (e.g., by peak or non-peak, weekday or weekend). In our application, we found that setting $n = 5$ and $s = 1$ works well.

Many existing anomaly grouping methods only group anomalies in space for each time period, and not across time, or across time only as a second step by clustering spatial subgraphs \cite{Pan:2013, Doraiswamy:2014}. 
Our approach is simple, fast, can be easily done online, and gives groupings across both space and time in one step.  

In Figure \ref{fig:nfl}, we highlight using ellipses the 5 anomaly events formed using this approach. This grouping in time and space is sensible and visually appealing. Some of these events may not be or are only partially due to the NFL game (e.g., event 2); there is no easy way to conclusively attribute the causes.

With the anomalies grouped into events, we can quantify the nature of each event in many ways. For example,  
\vspace{-2mm}
\begin{itemize}
\item Distribution in time and space: measure of ``center" and ``spread" of the event in time and space; characteristics of roads and time periods involved;
\item Distribution of anomaly values: maximum, median and minimum of $\tilde{A}$; an overall measure of ``seriousness'' such as $\sum |\tilde A|$, or some topological volume measure similar to those used in \cite{Doraiswamy:2014};
\item Lifecycle of anomalies: time span, speed, and acceleration of event ``birth'' leading up to peak time of event and ``death'' after the peak;
\item Spreading of anomalies: structure and compactness of the subgraph such as average path length and node degree; empirical propagation birth and death rate.
\end{itemize}
\vspace{-2mm}
These quantifications make it possible to cluster events, find similar past events, or compare events or clusters of events across time and regions, or before and after certain change in the traffic system. This provides an easily interpretable framework for non-experts such as urban planners to monitor and analyze urban traffic. 

For instance, the impacts of large urban events such as sports games on the urban traffic can be analyzed, e.g., where and when attendees come from or leave to, or potential ways to proactively alleviate traffic problems e.g. by providing extra public transportation. Analysis of event clusters can also enable urban planners to identify high-impact infrastructure projects that would improve traffic flow, such as identifying roads that are congested on a regular basis and increasing their capacity or public transportation, or prioritizing road work or maintenance resources to roads that experience the strongest congestion. 

\section{Conclusion}
We use anonymized and aggregate cellular location data to measure urban mobility, and propose a scheme that decomposes traffic into expected traffic and anomalies caused by various reasons. This decomposition is done at fine resolutions in time (hourly) and space (for each road).  Our approach is fast, online, and highly robust, even when anomalies are loosely correlated across time, and when traffic is heterogeneous and hard to parametrize.   

We adapt stable principal component pursuit to time series data and output normalized and interpretable quantifications of traffic expectations and anomalies that is comparable across time and roads.  We then group the road-level anomalies in time and space into anomaly events using a graph expansion procedure. We propose several ways to quantify the anomaly events.

Given these quantifications of traffic expectations and anomalies, we can locate budding anomalies as early in time and as precisely in space as possible for timely alerting or re-routing. The quantifications also provide an interpretable framework for urban planners to visualize and analyze urban traffic. We show that our system can help estimate the traffic impact related to major urban events, and can inform mitigation strategies; we suggest potential opportunities to use our outputs to identify high-impact infrastructure projects. 

\section{Acknowledgments}

We would like to thank Raif Rustamov (AT\&T Labs) for the considerable help with the location data and their probabilistic route assignments, as well as DeDe Paul for supporting this project as the director of AT\&T Labs' Statistics Research Department. We would also like to thank the d4gx reviewers for their helpful comments. 

\nocite{*}
\bibliographystyle{abbrv}
\bibliography{sigproc}

\end{document}